\pdfoutput=1
\documentclass[11pt]{article}
\usepackage[final]{acl}
\usepackage{times}
\usepackage{latexsym}
\usepackage[table]{xcolor}
\usepackage{colortbl}

\usepackage[T1]{fontenc}

\usepackage[utf8]{inputenc}

\usepackage{microtype}

\usepackage{inconsolata}

\usepackage{graphicx}
\usepackage{booktabs}

\usepackage{multirow}

\definecolor{glight}{RGB}{232,245,233}
\definecolor{gmed}{RGB}{165,214,167}
\definecolor{gdark}{RGB}{76,175,80}

\definecolor{rlight}{RGB}{255,235,238}
\definecolor{rmed}{RGB}{239,154,154}
\definecolor{rdark}{RGB}{229,115,115}

\definecolor{headergray}{RGB}{242,242,242}

\title{COILD: An Indic-Centric Parallel Corpus and Benchmark for\\
Machine Translation Across Indian Languages}

\author{\\
Kshetrimayum Boynao Singh$^{1}$,
Nitin Kumar Mishra$^{2}$,
Palash Pratim Dutta$^{3}$,
Atai Waris Khan$^{4}$,\\
Aparna Kaushik$^{6}$,
Avinash Kumar$^{1}$,
Deeksha$^{5}$,
Deepak Kumar$^{1}$,
Saroj Kumar Jha$^{1}$,\\
Saloka Sengupta$^{1}$,
Anansa Roy$^{1}$,
Umalatha Kannoth$^{5}$,
Saifulla Samar$^{3}$,
Meena Sharma$^{4}$,\\
Manpreet Kaur$^{6}$,
Jyoti Sharma$^{2}$,
Ashwini Vaidya$^{2}$,
Muralikrishna SN$^{5}$,
Md Shad Akhtar$^{4}$,\\
Poonam Bansal$^{6}$,
Amita Dev$^{6}$,
Sanasam Ranbir Singh$^{3}$,
Samit Bhattacharya$^{3}$,\\
Tanmoy Chakraborty$^{2}$,
Asif Ekbal$^{1}$
\\[0.1em]
{\normalsize
$^{1}$IIT Patna
\quad $^{2}$IIT Delhi
\quad $^{3}$IIT Guwahati
\quad $^{4}$IIIT Delhi
\quad $^{5}$MIT-MAHE
\quad $^{6}$IGDTUW
}
}

\begin{document}
\maketitle
\begin{abstract}

Machine translation (MT) for Indian languages remains constrained by the limited availability of high-quality, Indic-centric parallel corpora and evaluation benchmarks. Existing multilingual resources are largely constructed from English-pivot content and often fail to capture the linguistic diversity, cultural complexity, and domain-specific characteristics of Indian languages. We present \textbf{COILD}, an Indic-centric parallel corpus comprising over 1.16 million human-translated and human-verified sentence pairs, covering 20 Indian language pairs across the Indo-Aryan, Dravidian, Tibeto-Burman and Austro-Asiatic language families. The corpus is built entirely from original Indian language sources collected from licensed repositories that spans across eight domains with direct real-world applicability. Furthermore we introduce a domain-centric benchmark comprising 2,000 expert-verified sentences to enable consistent multilingual and cross-lingual evaluation across Indian language pairs. To validate the effectiveness of COILD, we fine-tune two representative multilingual neural machine translation models, IndicTrans2-Distilled and NLLB-200. Experimental results demonstrate consistent improvements across language pairs, domains, automatic evaluation metrics, and human evaluation, highlighting the effectiveness of high-quality Indic-centric supervision. COILD provides a valuable training and evaluation resource\footnote{\url{https://huggingface.co/datasets/coild-dataset/COILD-MT-Corpus}\\
\url{https://huggingface.co/COIL-D/models}\\
Data request: \url{https://forms.gle/xLG557PEizFGuL4z9}} for advancing multilingual machine translation and future multilingual language models for Indian languages.

\end{abstract}

\section{Introduction}

India is home to one of the world’s most linguistically diverse populations, with 22 languages listed in the Eighth Schedule of its Constitution, representing several language families, including Indo-Aryan, Dravidian, Tibeto-Burmese and Austro-Asiatic. These languages are spoken by more than a billion people and are increasingly becoming central to digital governance, education, healthcare, judiciary, and public communication. As multilingual applications continue to expand, high-quality machine translation (MT) systems have become essential for enabling inclusive access to digital services across diverse linguistic communities.

Despite recent progress in multilingual neural machine translation, the development of MT systems for Indian languages remains constrained by the quality and nature of available parallel corpora. Existing large-scale multilingual datasets, including Samanantar \citep{ramesh-etal-2022-samanantar}, NLLB \citep{nllbteam2022languageleftbehindscaling}, and BPCC \citep{gala2023indictrans}, are primarily constructed using English as the central pivot language. Although these resources provide substantial multilingual coverage, their construction strategy inherently introduces an English-centric perspective. Most Indic--Indic sentence pairs are either extracted through English or aligned using English as an intermediate representation. Consequently, they often fail to preserve the lexical, morphological, syntactic, and cultural characteristics naturally shared among Indian languages.

Beyond the issue of English pivoting, another important limitation lies in the origin of the source content itself. A significant portion of existing multilingual corpora is collected from globally sourced or web-mined content that is not originally written in Indian languages. Such datasets frequently under-represent Indian linguistic styles, domain-specific terminology, administrative expressions, and culturally grounded discourse. As a result, translation systems trained and evaluated on these corpora may achieve competitive automatic evaluation scores yet fail to generalise effectively to real-world Indian-language applications.

\begin{figure*}[t]
    \centering
    \includegraphics[width=0.95\textwidth, height=0.60\textheight, keepaspectratio]{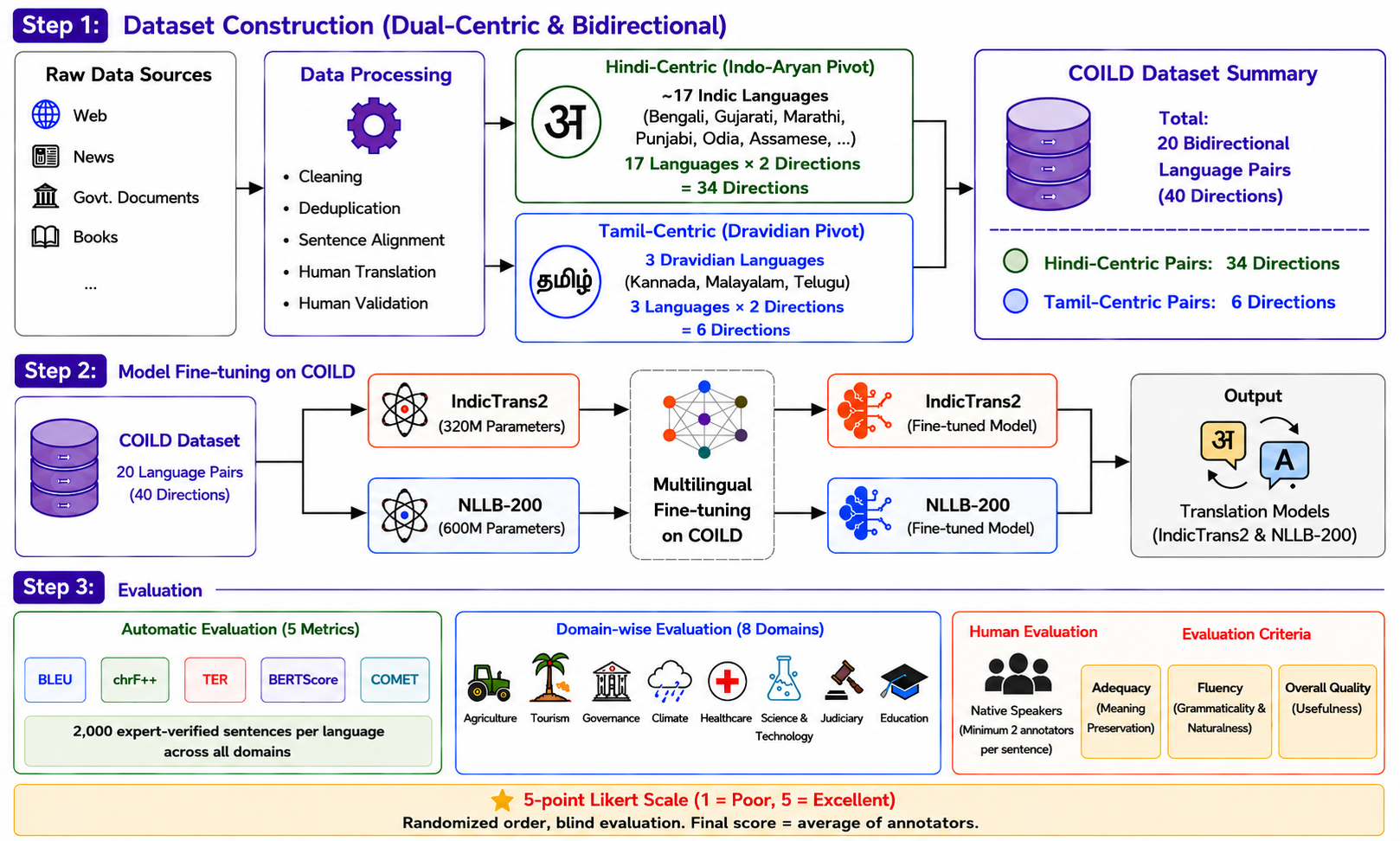}
    % \caption{Overview of the proposed COILD framework. The workflow consists of (1) dual-centric dataset construction, (2) multilingual fine-tuning of IndicTrans2 and NLLB-200, and (3) automatic and human evaluation.}

    \caption{Overview of the COILD framework and evaluation pipeline. The workflow consists of three stages: (1) \textit{dataset construction}, using a dual-centric and bidirectional strategy with Hindi as the Indo-Aryan pivot and Tamil as the Dravidian pivot, resulting in 20 language pairs and 40 translation directions; (2) \textit{model fine-tuning}, where IndicTrans2 and NLLB-200 are fine-tuned on the COILD parallel corpus; and (3) \textit{evaluation}, comprising automatic evaluation using five MT metrics, domain-wise evaluation across eight application domains, and human evaluation based on translation adequacy, fluency, and overall quality using a 5-point Likert scale.}
    \label{fig:framework}
\end{figure*}

This representational asymmetry also affects machine translation evaluation. Widely adopted multilingual benchmarks such as FLORES-200 and IN22 are valuable resources for standardised evaluation; however, they are largely derived from English-pivot source material and do not adequately represent the linguistic diversity and applied domains encountered in India. Consequently, evaluating Indic machine translation systems with these benchmarks fails to adequately represent the sociolinguistic realities of Indian language contexts, particularly in domains such as governance, education, healthcare, judiciary, tourism, and public administration.

To address these limitations, we introduce \textbf{COILD} (\textbf{C}entre of \textbf{I}ndian \textbf{L}anguage \textbf{D}ata)\footnote{\url{https://coild-bhashini.github.io/}}, an Indic-centric parallel corpus constructed entirely from content originally written in Indian languages. Unlike existing resources, COILD adopts a dual-pivot strategy in which Hindi serves as the pivot language for Indo-Aryan, Tibeto-Burman and Austro-Asiatic languages, and Tamil serves as the pivot for Dravidian languages. Source documents are collected from licensed Indian resources, including government publications, newspapers, educational materials, books and works by authors, publications, and bloggers who volunteered to provide source materials, as well as other publicly available repositories. Each sentence undergoes rigorous preprocessing, human translation, expert verification, and terminology validation before inclusion in the corpus. The resulting dataset consists exclusively of human-translated and human-verified parallel sentences, eliminating the need for web mining, automatic sentence alignment, or back-translation. Recognising that practical machine translation performance varies significantly across application domains, we further construct a domain-balanced evaluation benchmark covering eight important sectors: Governance, Education, Healthcare, Judiciary, Science \& Technology, Tourism, Climate and Agriculture. The benchmark is held out entirely from training and enables systematic evaluation of translation quality across both language pairs and domains, providing a more realistic assessment of MT systems intended for Indian applications.

We evaluate the effectiveness of COILD by fine-tuning two representative multilingual translation models, IndicTrans2 and NLLB-200. Experimental results demonstrate consistent improvements across nearly all translation directions evaluated, highlighting the importance of high-quality Indic-centric supervision. Furthermore, our analysis shows that constructing parallel data from typologically appropriate Indian language pivots substantially improves translation quality compared to conventional English-centric approaches, while our human evaluation reveals important limitations of widely used automatic evaluation metrics for Indian languages.

Our main contributions are summarised as follows:

\begin{itemize}
    \item We introduce \textbf{COILD}, a large-scale Indic-centric parallel corpus comprising over 1.16 million human-translated and human-verified sentence pairs and approximately 44.79 million tokens across source and target sides, constructed from original Indian-language sources rather than English-mediated content.

    \item We propose a dual-pivot corpus construction strategy that employs Hindi for Indo-Aryan, Tibeto-Burman and Austro-Asiatic languages and Tamil for Dravidian languages, preserving linguistic characteristics shared within each language family.

    \item We develop a domain-specific benchmark comprising expert-verified translations across eight real-world domains, enabling more realistic evaluation of machine translation systems for Indian applications.

    \item We demonstrate that fine-tuning state-of-the-art multilingual translation models on COILD consistently improves translation quality across multiple Indian language pairs, while providing an extensive analysis of pivot selection and the reliability of automatic evaluation metrics for Indic machine translation.
\end{itemize}

\section{Related Work}
\label{sec:related_work}

\subsection{Parallel Corpora for Indian Languages}

The availability of high-quality parallel corpora has been one of the primary factors driving progress in machine translation for Indian languages. Early efforts focused on manually curated bilingual resources. The Indian Languages Corpora Initiative (ILCI) \citep{jha2010tdil} introduced one of the first large-scale translation datasets for Indian languages, where Hindi served as the primary source language for creating parallel corpora across multiple Indic languages. Although the corpus provided high-quality human translations, its relatively limited scale restricted its applicability for training modern neural machine translation (NMT) systems.

With the advent of multilingual neural machine translation, research shifted towards constructing large-scale parallel corpora through automatic mining and alignment. Samanantar \citep{ramesh-etal-2022-samanantar} significantly expanded the availability of parallel data by mining approximately 49 million sentence pairs covering eleven Indian languages. Subsequently, No Language Left Behind (NLLB) \citep{nllbteam2022languageleftbehindscaling} extended multilingual parallel data collection to more than 200 languages using large-scale web mining and automatic sentence alignment. BPCC\footnote{\url{https://huggingface.co/datasets/ai4bharat/BPCC}} \citep{gala2023indictrans} further increased the coverage of Indian languages by introducing over 230 million sentence pairs spanning all 22 scheduled Indian languages.

These resources have substantially improved multilingual translation research; however, they share several common characteristics. First, most are constructed using English as the central source or alignment language, resulting in many Indic--Indic sentence pairs being derived indirectly rather than translated directly between Indian languages. Second, a considerable portion of the data originates from web-mined multilingual content, where sentence alignment errors, translation inconsistencies, and noisy text inevitably affect corpus quality \citep{kreutzer2022quality}. Finally, the source material itself is often not originally written in Indian languages, resulting in limited representation of Indian linguistic expressions, administrative terminology, cultural references, and domain-specific discourse. In contrast, COILD focuses on constructing an Indic-centric corpus from content originally written in Indian languages, including eight domains. Rather than relying on automatic mining or sentence alignment, every sentence pair is produced through human translation and expert verification, ensuring high linguistic quality while preserving the natural characteristics of Indian languages.

\subsection{Machine Translation Benchmarks}

Reliable evaluation benchmarks are equally important for measuring translation quality. FLORES-101 \citep{goyal2022flores101} and FLORES-200 \citep{nllbteam2022languageleftbehindscaling} established multilingual evaluation benchmarks covering hundreds of languages through professionally translated reference sets. NTREX-128 \citep{federmann2022ntrex128} further extended multilingual evaluation using news-domain sentences, while IN22 \citep{gala2023indictrans} introduced an evaluation benchmark covering all scheduled Indian languages.

Although these benchmarks provide standardised evaluation protocols, they are largely constructed from English-origin source sentences and primarily focus on measuring translation performance under generalised multilingual settings. Consequently, they often lack the linguistic diversity, cultural complexity, and domain-specific content encountered in real-world Indian applications. Furthermore, existing benchmarks generally provide limited support for systematic domain-wise evaluation, making analysis difficult. We introduce eight real-world application domains, namely Agriculture, Climate, Education, Governance, Healthcare, Judiciary, Science \& Technology, and Tourism, providing diverse linguistic coverage for multilingual translation. To address these limitations, COILD introduces an Indic-centric benchmark comprising 2,000 expert-verified sentences distributed across eight carefully selected application domains. The benchmark is designed to support consistent evaluation across multiple Indian language pairs and to serve as a reusable resource for future multilingual and cross-lingual machine translation systems.

\subsection{Multilingual Machine Translation for Indian Languages}

Recent advances in multilingual neural machine translation have significantly improved translation quality for low-resource languages. IndicTrans2 \citep{gala2023indictrans} introduced a many-to-many multilingual translation framework specifically optimised for all scheduled Indian languages, demonstrating substantial improvements over previous multilingual systems. NLLB-200 \citep{nllbteam2022languageleftbehindscaling} further showed that massively multilingual models can achieve competitive performance across hundreds of languages using large-scale multilingual supervision. Other studies have explored direct Indic-to-Indic neural machine translation \citep{das2024multilingual}, multilingual transfer learning, and language-family-aware training strategies to improve translation between related Indian languages.

Despite these architectural advancements, the performance of multilingual MT systems continues to depend heavily on the availability of high-quality parallel data. Improvements in model architectures alone cannot compensate for limitations in the underlying training resources. High-quality human-translated corpora remain scarce for many Indian language pairs, while realistic evaluation benchmarks that reflect Indian domains are even scarcer.

\subsection{Our Contribution}

COILD is designed as an Indic-centric resource rather than an English-centric multilingual corpus. The proposed dataset contains more than 1.16 million human-translated and human-verified sentence pairs covering 20 Indian language pairs across Indo-Aryan, Tibeto-Burman, Austro-Asiatic and Dravidian language families. In addition, we introduce a domain-centric benchmark consisting of 2,000 expert-verified sentences spanning eight real-world domains, enabling consistent evaluation across multiple Indian language pairs and facilitating future multilingual and cross-lingual MT research. Finally, we demonstrate the effectiveness of the proposed corpus by fine-tuning two representative multilingual MT systems, IndicTrans2 and NLLB-200, showing that high-quality Indic-centric supervision consistently improves translation performance.

\begin{table*}[t]
\centering
\small
\setlength{\tabcolsep}{5pt}

\begin{tabular}{ll c cc cc}
\toprule
\multirow{2}{*}{\textbf{Source}} &
\multirow{2}{*}{\textbf{Target}} &
\multirow{2}{*}{\textbf{Sentence Pairs}} &
\multicolumn{2}{c}{\textbf{Token Count}} &
\multicolumn{2}{c}{\textbf{Vocabulary}} \\
\cmidrule(lr){4-5}\cmidrule(lr){6-7}
& & &
\textbf{Source} &
\textbf{Target} &
\textbf{Source} &
\textbf{Target} \\
\midrule

Hindi & Assamese & 94,872 & 19,86,178 & 15,90,680 & 76,956 & 1,18,170 \\
Hindi & Bengali & 73,066 & 14,89,733 & 11,76,804 & 58,937 & 79,616 \\
Hindi & Bodo & 47,599 & 9,88,036 & 7,62,802 & 47,698 & 74,469 \\
Hindi & Dogri & 58,896 & 11,93,781 & 12,09,040 & 53,385 & 56,814 \\
Hindi & Gujarati & 96,612 & 20,22,924 & 16,84,920 & 78,395 & 1,15,035 \\
Hindi & Kashmiri & 58,703 & 11,42,064 & 11,25,761 & 50,943 & 1,01,103 \\
Hindi & Konkani & 49,631 & 10,16,959 & 7,66,060 & 47,298 & 93,018 \\
Hindi & Maithili & 93,748 & 19,60,050 & 18,53,832 & 73,530 & 86,706 \\
Hindi & Manipuri & 52,494 & 10,63,191 & 8,36,037 & 50,875 & 94,357 \\
Hindi & Marathi & 60,288 & 12,62,360 & 9,62,931 & 56,671 & 96,449 \\
Hindi & Nepali & 61,643 & 13,08,102 & 10,14,659 & 56,636 & 1,05,664 \\
Hindi & Odia & 52,733 & 11,01,517 & 8,88,246 & 50,359 & 67,619 \\
Hindi & Punjabi & 98,163 & 20,63,467 & 20,71,497 & 79,268 & 69,471 \\
Hindi & Santali & 20,603 & 3,92,518 & 3,80,606 & 25,736 & 42,599 \\
Hindi & Sindhi & 43,663 & 8,69,067 & 8,75,327 & 42,052 & 64,937 \\
Hindi & Telugu & 64,484 & 13,70,263 & 9,51,100 & 60,365 & 95,530 \\
Hindi & Urdu & 84,205 & 17,77,457 & 18,84,100 & 69,271 & 82,278 \\

\midrule

Tamil & Kannada & 30,274 & 3,89,468 & 3,85,899 & 79,173 & 72,772 \\
Tamil & Malayalam & 23,593 & 2,92,095 & 2,75,648 & 67,797 & 69,650 \\
Tamil & Telugu & 16,034 & 1,93,271 & 2,08,372 & 50,373 & 37,707 \\

\bottomrule
\end{tabular}

\caption{Statistics of the COILD corpus. The corpus contains Hindi-centric (Hindi $\rightarrow$ 17 target languages) and Tamil-centric (Tamil $\rightarrow$ 3 target languages) parallel corpora. We report the number of aligned sentence pairs, total token counts, and vocabulary sizes for the source and target sides. The COILD corpus contains approximately 1.16 million human-translated sentence pairs, comprising 44.79 million tokens across the source and target sides.}
\label{tab:corpus}
\end{table*}

\section{The COILD Corpus}
\label{sec:corpus}

The primary objective of COILD\footnote{\url{https://huggingface.co/datasets/coild-dataset/COILD-MT-Corpus}} is to construct a large-scale, high-quality, Indic-centric parallel corpus that accurately reflects the linguistic diversity, cultural context, and domain-specific characteristics of Indian languages. Unlike existing multilingual resources that are predominantly derived from English-centric or automatically mined content, every stage of COILD, from source acquisition to translation and quality assurance, is designed around original Indian language content and human verification. Figure~\ref{fig:framework} presents an overview of the complete corpus construction pipeline.

\subsection{Corpus Design Principles}

The design of COILD is guided by four fundamental principles.

\begin{itemize}
    \item \textbf{Indic-centric source construction:} All source documents originate from Indian language resources rather than translated English documents, ensuring that the linguistic and cultural characteristics of Indian languages are preserved throughout the corpus.

    \item \textbf{Human-centric translation:} Every sentence in the corpus is translated manually by qualified translators without using machine translation systems, followed by multiple rounds of linguistic verification.

    \item \textbf{Domain diversity:} The corpus is collected from multiple real-world domains to improve its applicability for practical machine translation systems.

    \item \textbf{Quality-first methodology:} Every stage of corpus creation incorporates human validation and quality control, ensuring that only expert-verified translations are included in the final release.
\end{itemize}

\subsection{Source Collection and Validation}

COILD is constructed from original Indian-language documents sourced from licensed and publicly available resources, including government publications, policy documents, educational materials, newspapers, books, and other domain-specific repositories. Appropriate copyright permissions or usage approvals were obtained wherever required before incorporating the data into the corpus.

Many source documents were available as scanned PDFs or image-based archives; therefore, Optical Character Recognition (OCR) \citep{CDEC-OCR} was applied to extract editable text. The extracted content was manually cleaned to correct OCR errors, remove formatting artefacts, normalise Unicode text, and segment documents into sentence-level units.

The processed sentences were subsequently uploaded to our in-house annotation platform, \textit{PostEditMe}\footnote{\url{https://posteditme.in/}}, where language experts performed monolingual validation. During this stage, reviewers corrected grammatical and spelling errors, removed duplicate, incomplete, and ambiguous sentences, filtered personally identifiable information, and ensured contextual completeness. Only linguistically well-formed and contextually independent sentences were retained for translation, providing high-quality source data for corpus construction.

\subsection{Translator Recruitment and Selection}

Producing high-quality multilingual translations requires qualified translators with strong proficiency in the relevant source and target languages. For the Hindi-centric language pairs, translators were required to be proficient in Hindi and the corresponding target language, including Assamese, Bengali, Bodo, Dogri, Gujarati, Kashmiri, Konkani, Maithili, Manipuri, Marathi, Nepali, Odia, Punjabi, Santali, Sindhi, Telugu and Urdu. For the Tamil-centric language pairs, translators were required to be proficient in Tamil and the corresponding target languages, namely Kannada, Malayalam, and Telugu. 

The recruitment was conducted through an open call distributed across diverse Indian states and language communities to minimise regional and dialectal bias. Finding suitable candidates proved challenging, particularly for low-resource languages, as qualified bilingual translators were relatively limited. In addition, some potential candidates, particularly from older generations, lacked the computer skills needed to use modern annotation platforms, further reducing the available pool of translators. 

Applicants were evaluated through a two-stage selection process to verify their language proficiency and professional experience. The first stage involved background verification, including assessment of educational background, relevant translation experience, and proficiency in the required languages. The second stage consisted of an expert-designed translation assessment. Candidate translations were manually evaluated for adequacy, fluency, grammatical correctness and consistency in terminology. Only candidates who successfully passed both stages were onboarded onto the annotation platform.

To maintain consistency across language pairs, all translators received detailed translation guidelines and participated in onboarding sessions covering translation policies, quality expectations, and platform usage. Throughout the annotation process, translators were explicitly prohibited from using existing machine translation systems or other automated translation tools. All translations were required to be produced manually based solely on the provided source sentences and the accompanying terminology and translation guidelines.

\subsection{Human Translation Workflow}

Translation was performed entirely through the PostEditMe annotation platform. Source sentences were assigned to each translator in batches of approximately 100. Every sentence was translated independently by a native speaker of the target language following the standardised translation protocol. The guidelines instructed translators to preserve the semantic meaning, discourse structure, register, and domain-specific terminology of the source sentence while producing natural target-language expressions. Literal word-by-word translation was discouraged whenever it affected fluency or readability. Each completed batch was automatically submitted for linguistic review immediately after annotation.

\subsection{Multi-stage Quality Assurance}
To ensure high translation quality, COILD employs a multi-stage human-verification pipeline. Unlike existing multilingual corpora that primarily rely on automatic filtering or sentence alignment, every translated sentence undergoes expert review before inclusion in the final corpus. Each translation batch is independently reviewed by in-house linguistic experts from participating consortium institutions, who are different from the original translators. The reviewers perform line-by-line verification based on four quality criteria: (i) semantic adequacy, (ii) linguistic fluency, (iii) terminology consistency, and (iv) grammatical correctness. Translation quality is assessed using a TER-based human post-editing process. Reviewers edit the submitted translations where necessary, and the platform automatically computes the edit rate between the original and revised versions. Batches with an edit rate below 20\% are accepted, while those exceeding the threshold are returned to the translators for revision. This review-revision cycle continues until the required quality standard is achieved. Finally, all accepted translations undergo domain-specific validation by subject experts to ensure the correctness of technical terminology, contextual consistency, and domain relevance before being incorporated into the final COILD corpus.

\subsection{Benchmark Construction}

\begin{figure}
    \centering
    \includegraphics[width=1\linewidth]{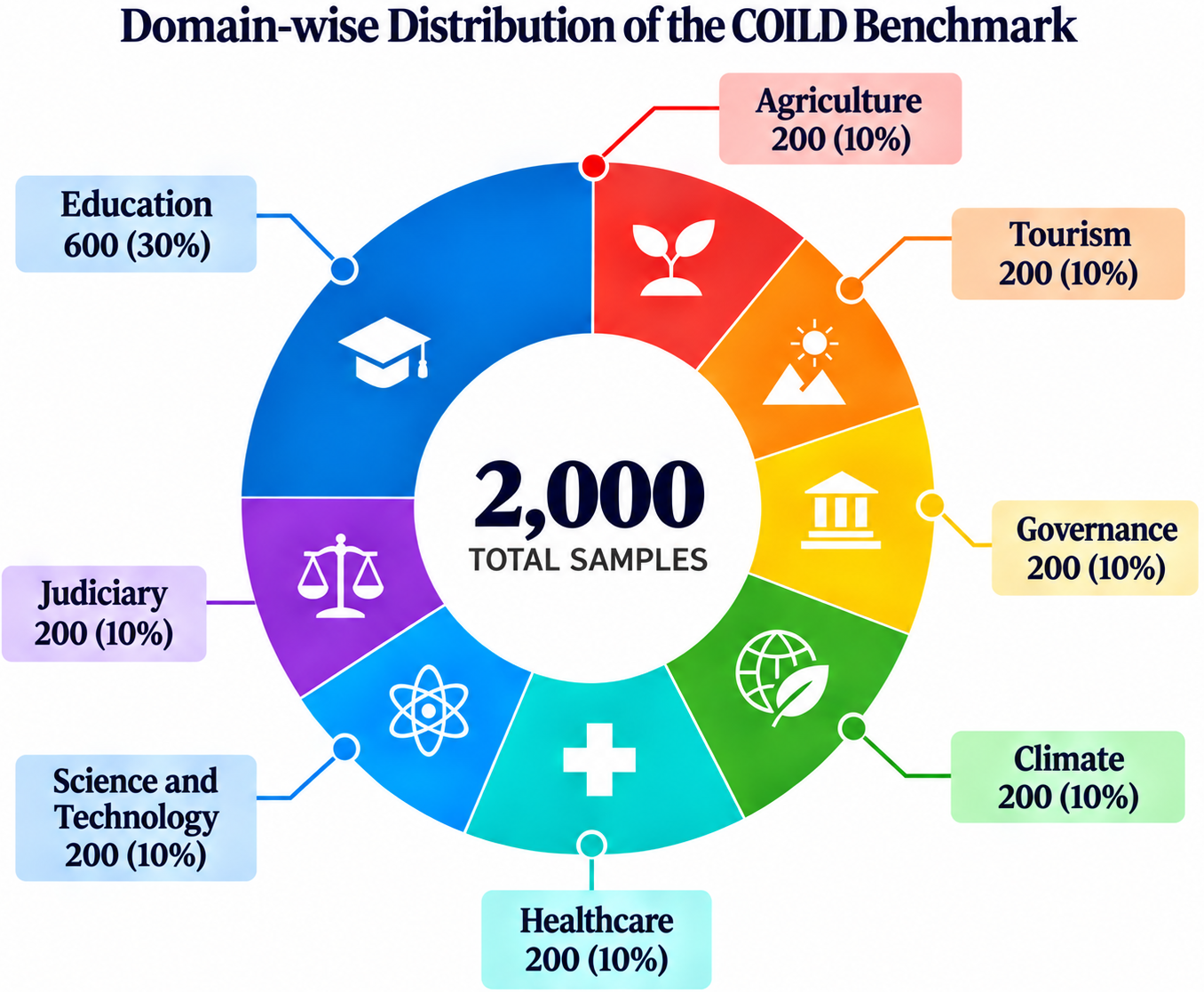}
    \caption{Domain-wise distribution of the COILD benchmark, comprising 2,000 Hindi source sentences across eight domains. The Education domain contains 600 sentences, while each of the remaining seven domains contains 200 sentences.}
    \label{fig:benchmark_domains}
\end{figure}

In addition to the training corpus, we construct a dedicated evaluation benchmark consisting of 2,000 expert-verified parallel sentences. Unlike the training corpus, benchmark creation is performed exclusively by in-house language experts from consortium institutions. The benchmark covers eight application domains, with the domain-wise sentence distribution summarised in Figure~\ref{fig:benchmark_domains}. Source sentences are carefully selected to represent varying levels of translation difficulty and domain complexity. Every benchmark sentence is independently translated by one language expert and subsequently verified by another expert who did not participate in the original translation, thereby minimising evaluator bias. All benchmark translations preserve the same sentence identifiers across all languages. Consequently, a given sentence identifier corresponds to exactly the same source sentence in every supported language: Assamese, Bengali, Bodo, Dogri, Gujarati, Kashmiri, Konkani, Maithili, Manipuri, Marathi, Nepali, Odia, Punjabi, Santali, Sindhi, Telugu and Urdu for the Hindi-centric pairs, and Kannada, Malayalam and Telugu for the Tamil-centric pairs. This aligned design enables not only conventional source-to-target evaluation but also systematic evaluation of arbitrary Indian language pairs without requiring additional reference creation. The resulting benchmark therefore provides a reusable evaluation resource for multilingual, many-to-many, and cross-lingual machine translation systems while enabling consistent domain-wise performance analysis across Indian languages.

\section{Experimental Setup}
\label{sec:experimental_setup}

The objective of our experiments is not to propose a new machine translation architecture, but rather to evaluate the effectiveness of the proposed \textbf{COILD} corpus as a high-quality training resource for multilingual machine translation. To this end, we fine-tune representative state-of-the-art multilingual neural machine translation (NMT) models using the proposed corpus and evaluate their performance on the held-out COILD benchmark. Our experiments are designed to answer the following research questions:
\begin{itemize}
    \item Does an Indic-centric parallel corpus improve multilingual machine translation performance?
    \item Can the proposed corpus consistently benefit models with different multilingual training paradigms?
    \item Does the proposed benchmark provide a reliable evaluation framework across multiple Indian language pairs and application domains?
\end{itemize}

\subsection{Models}

To validate the generality of COILD, we select two complementary multilingual NMT models that represent different design philosophies.

\textbf{IndicTrans2-Distilled} is a multilingual translation model specifically developed for Indian languages. It is trained using multilingual supervision across Indic language pairs while maintaining a lightweight architecture suitable for efficient deployment. Since the model is optimised for Indic multilingual translation, it provides an appropriate baseline for evaluating whether COILD offers additional benefits beyond existing Indic-focused training resources.

We further evaluate \textbf{NLLB-200}, a massively multilingual translation model supporting over 200 languages. Unlike IndicTrans2, NLLB is trained on large-scale multilingual data collected from diverse languages and domains. Evaluating COILD on NLLB enables us to investigate whether an Indic-centric corpus can improve translation quality even for a general-purpose multilingual model.

Together, these models allow us to assess the effectiveness of COILD across both an Indic-specialised translation model and a large-scale multilingual translation system. For both models, we initialise from the official pretrained checkpoints and fine-tune them using the proposed COILD corpus.

\subsection{Training Data}

All experiments are conducted using the COILD training corpus described in Section~\ref{sec:corpus}. The corpus consists of more than 1.16 million human-translated and human-verified sentence pairs covering 20 Indian language pairs across Indo-Aryan, Tibeto-Burman, Austro-Asiatic and Dravidian language families. The data spans eight real-world application domains, namely Agriculture, Climate, Education, Governance, Healthcare, Judiciary, Science \& Technology, and Tourism, providing diverse linguistic coverage for multilingual translation.

Each language pair is fine-tuned independently from the corresponding pretrained checkpoint, isolating the contribution of COILD supervision for each translation direction while preserving the original multilingual representations learned during pre-training.

\subsection{Implementation Details}

Both models are fine-tuned using their officially recommended optimisation settings while retaining their original multilingual vocabularies and tokenisation strategies. Mixed-precision (FP16) training is employed to improve computational efficiency without affecting translation quality. Model selection is based on validation \textit{chrF++}, and the checkpoint that achieves the highest validation performance is used for all subsequent evaluations. The complete hyperparameter configuration is presented in Table~\ref{tab:hyperparameters}.

\begin{table}[t]
\centering
\small
\setlength{\tabcolsep}{4pt}
\renewcommand{\arraystretch}{1.05}
\begin{tabular}{lcc}
\toprule
\textbf{Hyperparameter} & \textbf{IndicTrans2} & \textbf{NLLB-200} \\
\midrule
Epochs & 5 & 5 \\
Learning Rate & $2\times10^{-5}$ & $2\times10^{-5}$ \\
LR Scheduler & Cosine & Linear \\
Warm-up & 1\% & 3\% \\
Optimizer & AdamW (8-bit) & AdamW (8-bit) \\
Effective Batch Size & 64 & 64 \\
Precision & bf16 & bf16 \\
\bottomrule
\end{tabular}
\caption{Hyperparameter configuration used for fine-tuning IndicTrans2 and NLLB}
\label{tab:hyperparameters}
\end{table}

\subsection{Evaluation Protocol}

We evaluate all models on the proposed COILD benchmark, which comprises 2,000 expert-verified sentences across eight application domains. Since the benchmark maintains identical sentence alignment across all supported languages, it enables consistent evaluation of multiple Indian language pairs and supports multilingual and cross-lingual machine translation evaluation.

To obtain a comprehensive assessment of translation quality, we report five widely adopted automatic evaluation metrics: BLEU~\citep{papineni2002bleu}, chrF++~\citep{popovic2017chrfpp}, TER~\citep{snover2006ter}, BERTScore~\citep{zhang2020bertscore}, and COMET~\citep{rei2020comet}. BLEU, chrF++, and TER primarily measure lexical overlap and edit distance, whereas BERTScore and COMET capture semantic similarity using pretrained multilingual encoders. Unless otherwise stated, translations are generated using beam-search decoding with a beam size of five.

In addition to overall translation quality, we report domain-wise performance across the eight benchmark domains to analyse how the proposed corpus generalises to different real-world application scenarios.

\begin{figure*}[h]
\centering

\includegraphics[width=1\textwidth]{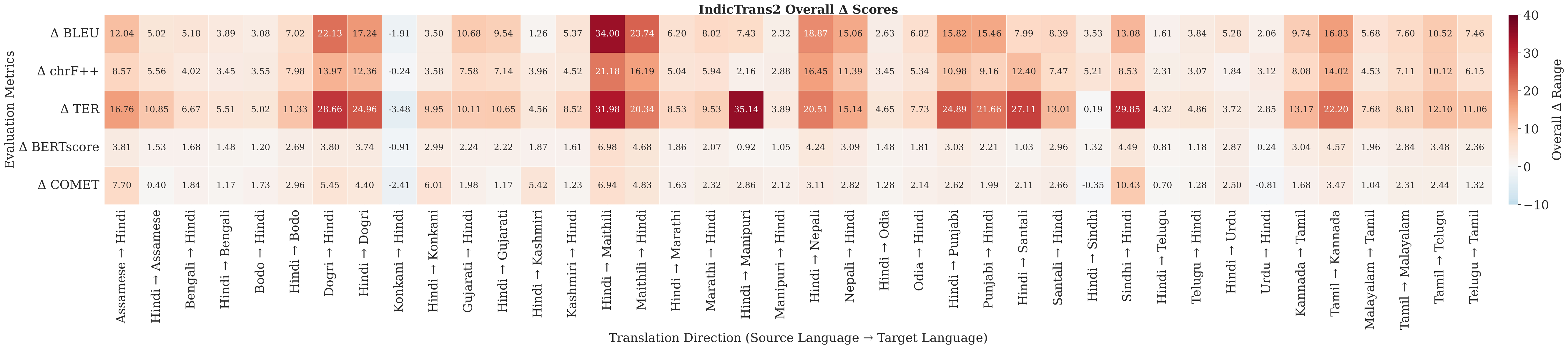}\\[2pt]
\includegraphics[width=1\textwidth]{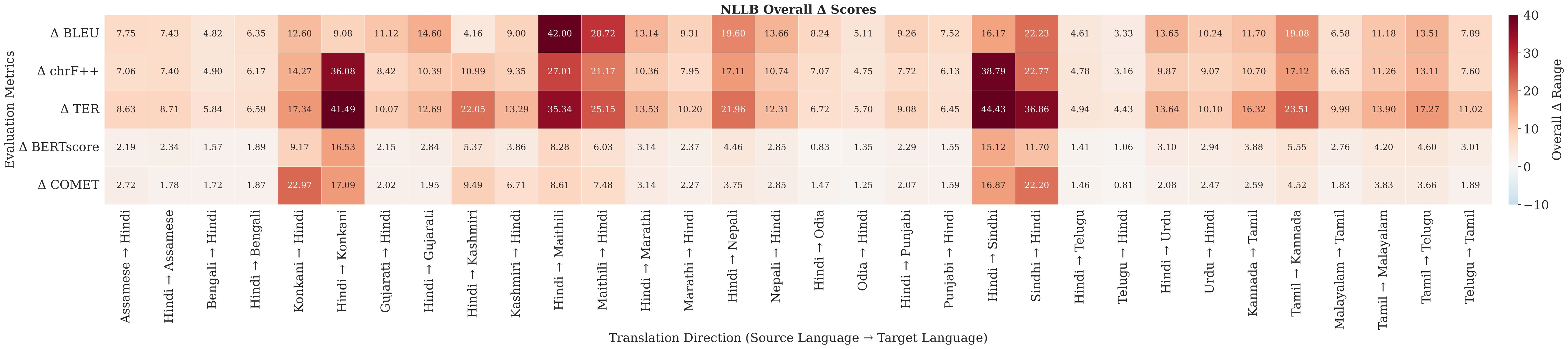}
\caption{Overall translation performance after fine-tuning on COILD for IndicTrans2-Distilled (top) and NLLB-200 (bottom). Heatmaps show the improvement over the corresponding pretrained baseline ($\Delta$ = Fine-tuned $-$ Baseline) across five evaluation metrics: BLEU, chrF++, TER, BERTScore, and COMET. Columns represent translation directions, rows denote evaluation metrics, and larger positive values indicate better performance achieved through fine-tuning on the COILD corpus.}
\label{fig:overall_delta}
\end{figure*}

\section{Results}
\label{sec:results}
This section evaluates the effectiveness of COILD as a high-quality Indic-centric parallel corpus for multilingual machine translation. Rather than proposing a new translation architecture, our objective is to investigate whether the proposed corpus provides effective supervision for existing multilingual MT models. To this end, we evaluate COILD from three complementary perspectives: (i) overall translation performance after fine-tuning, (ii) domain-wise generalisation using the proposed benchmark, and (iii) human evaluation to verify the quality of translated outputs.

\subsection{Overall Translation Performance}

Figure~\ref{fig:overall_delta} shows the overall translation performance after finetuning for both the model, Figure~\ref{fig:overall_baseline} and Figure~\ref{fig:overall_finetune} in the appendix section, present the translation performance of IndicTrans2-Distilled and NLLB-200 before and after fine-tuning on COILD across all supported translation directions.

Overall, fine-tuning on COILD consistently improves translation quality for both multilingual models across nearly all language pairs. Improvements are observed in both Indo-Aryan and Dravidian languages, indicating that the proposed corpus provides effective multilingual supervision regardless of the underlying model architecture. While IndicTrans2-Distilled already possesses strong Indic language representations, it continues to benefit substantially from COILD, demonstrating that high-quality human-translated data remains valuable even for Indic-specialised models. Similarly, NLLB-200, despite being pretrained on hundreds of languages, also shows consistent improvements after fine-tuning, suggesting that carefully curated Indic-centric supervision complements large-scale multilingual pretraining.

The largest improvements are generally observed for relatively low-resource language pairs and languages with comparatively limited parallel data, whereas language pairs with stronger pretrained baselines exhibit smaller but consistently positive improvements. This behaviour indicates that COILD effectively fills existing data gaps while simultaneously refining translation quality for well-supported languages.

\subsection{Domain-wise Performance Analysis}

One of the primary objectives of COILD is to provide broad domain coverage representative of real-world Indian language applications. Figure~\ref{fig:domain-bleu} visualises the domain-wise change in BLEU after fine-tuning on COILD. Each row corresponds to one benchmark domain, each column represents a translation direction, and each cell reports the BLEU improvement over the corresponding pretrained baseline. Darker shades indicate larger improvements, whereas blue cells represent performance degradation.

Across both IndicTrans2-Distilled and NLLB-200, positive improvements are consistently observed across all eight benchmark domains, demonstrating that the benefits of COILD generalise beyond a single application scenario. Although the magnitude of improvement varies across domains and language pairs, no domain exhibits systematic degradation after fine-tuning.

Positive gains are observed across all eight benchmark domains for both IndicTrans2 and NLLB, indicating that the effectiveness of COILD is not restricted to a specific application area. Although the magnitude of improvement varies across translation directions, every domain contains multiple language pairs showing significant BLEU improvement.

Among the evaluated domains, Governance, Judiciary, and Science \& Technology show the most substantial improvements for both models. These domains typically contain specialised terminology, longer sentence structures, and formal linguistic expressions, making them considerably more challenging for multilingual translation systems. Consistent improvements suggest that the domain-balanced construction of COILD successfully captures these linguistic characteristics and provides high-quality supervision for technical translation.

Education also demonstrates stable improvements across nearly all language pairs. Since this domain contains the largest number of benchmark sentences, the observed performance indicates that COILD generalises well across diverse educational topics and writing styles. Similarly, Agriculture, Tourism, Climate, and Healthcare consistently benefit from fine-tuning, although the magnitude of improvement is comparatively smaller for certain language directions. Overall, the domain-wise analysis shows that COILD improves translation quality and robustness across diverse real-world domains.

\subsection{Human Evaluation}
While automatic evaluation metrics provide a quantitative measure of translation quality, they can't fully capture linguistic naturalness and semantic correctness. To complement the automatic evaluation, the translations generated by the baseline and fine-tuned models are independently assessed by two native-language annotators for each evaluated language pair and translation direction. The evaluation considers two complementary criteria: \textit{Adequacy}, which measures how accurately the translation preserves the source meaning, and \textit{Fluency}, which measures the grammatical correctness and naturalness of the translated sentence. Each criterion is rated on a 5-point Likert scale (1 = Worst, 2 = Poor, 3 = Average, 4 = Good, 5 = Excellent), and the final score for each criterion is calculated as the average of the ratings assigned by the two annotators.

Table~\ref{tab:human_eval} presents the human evaluation results for the baseline and fine-tuned versions of IndicTrans2 and NLLB. Across most language pairs, fine-tuning on COILD yields consistent improvements in Adequacy and Fluency over the pretrained baselines. The improvements are particularly evident for several low-resource translation directions, where human evaluators observe more faithful semantic preservation and substantially improved readability after fine-tuning.

The greatest improvements are observed for several low-resource translation directions. For IndicTrans2, Urdu$\rightarrow$Hindi achieves the largest gain, improving by 2.35 points in Adequacy and 2.32 points in Fluency, followed by Hindi$\rightarrow$Maithili (1.46/1.49) and Maithili$\rightarrow$Hindi (1.27/1.30). NLLB shows similar trends across its supported language pairs. Larger gains are obtained for Hindi$\rightarrow$Konkani (2.13/1.94), Hindi$\rightarrow$Sindhi (2.41/2.25), Hindi$\rightarrow$Urdu (1.33/1.29), and Tamil$\rightarrow$Kannada (1.26/0.57), indicating that COILD also improves translation quality across different multilingual architectures.

For higher-resource language pairs such as Hindi--Gujarati, Hindi--Marathi, Hindi--Punjabi, and Hindi--Nepali, the improvements are generally smaller but remain consistently positive, indicating that COILD contributes to refining translation quality even when strong pretrained representations already exist. A small number of language directions exhibit modest decreases in human evaluation. However, these cases are limited and do not represent a consistent degradation trend across either model.

The agreement between automatic evaluation metrics and human judgments further validates the effectiveness of COILD. Language pairs that achieve larger improvements in BLEU and chrF++ also tend to receive higher adequacy and fluency scores, suggesting that the observed automatic metric gains correspond to genuine improvements in translation quality rather than metric-specific optimisation.

\section{Discussion}
\label{sec:discussion}

The experimental results demonstrate that the effectiveness of multilingual machine translation depends not only on model architecture but also on the quality of the underlying training data. Fine-tuning both IndicTrans2-Distilled and NLLB-200 on COILD consistently improves translation quality across most language pairs, indicating that the proposed corpus provides high-quality multilingual supervision irrespective of the pretrained model. The consistent improvements across two fundamentally different multilingual models suggest that COILD generalises well and is not tailored to a specific architecture.

The domain-wise evaluation further highlights the importance of constructing domain-balanced training resources. Improvements are observed across all eight application domains, with particularly strong gains in Governance, Judiciary, and Science \& Technology, where accurate translation of domain-specific terminology is essential. These findings indicate that incorporating diverse, high-quality Indian-language content enables models to generalise better across practical deployment scenarios. Human evaluation further supports the effectiveness of the proposed corpus. Independent assessments by two native-language annotators for each translation direction consistently report improvements in adequacy, fluency, and overall translation quality, aligning well with the automatic evaluation metrics. This agreement suggests that the observed improvements reflect genuine enhancements in translation quality rather than metric-specific optimisation.

We intentionally evaluate COILD using established multilingual NMT models instead of large language models (LLMs). Our goal is to isolate the contribution of the proposed dataset by using widely adopted, reproducible translation baselines. Nevertheless, COILD is not limited to conventional NMT. The high-quality, human-verified, and domain-balanced nature of the corpus makes it equally valuable for future multilingual LLM pretraining, instruction tuning, and domain adaptation for Indian languages.

Overall, COILD demonstrates that carefully curated Indic-centric parallel data remains a critical factor for advancing multilingual machine translation. We believe that the proposed corpus and benchmark will provide a strong foundation for future research on both neural machine translation and multilingual large language models for Indian languages.

\section{Conclusion}
\label{sec:conclusion}

In this paper, we presented \textbf{COILD}, a large-scale, human-translated and human-verified Indic-centric parallel corpus comprising over 1.16 million sentence pairs covering 20 Indian language pairs across eight real-world domains. We also introduced a domain-centric benchmark comprising 2,000 expert-verified sentences to enable consistent evaluation across multilingual and cross-lingual Indic machine translation tasks. To validate the proposed resource, we fine-tuned two representative multilingual translation models, IndicTrans2-Distilled and NLLB-200. Experimental results demonstrate consistent improvements across language pairs, application domains, automatic evaluation metrics, and human evaluation, highlighting the effectiveness of high-quality Indic-centric supervision for multilingual machine translation. These findings demonstrate the value of COILD as both a reliable training corpus and a reusable evaluation benchmark for Indian languages. Beyond neural machine translation, COILD can support future research in multilingual large language model (LLM) pretraining, instruction tuning, domain adaptation, and many-to-many Indic language translation. Overall, COILD provides a valuable resource for advancing multilingual NLP research and enabling more inclusive language technologies across India's diverse linguistic landscape.

\section*{Limitations}

Although COILD provides a large-scale, high-quality Indic-centric parallel corpus and benchmark, several limitations remain. First, the corpus currently covers 20 Indian language pairs across eight application domains and therefore does not represent the full linguistic diversity of India, including many regional dialects and minority languages. Second, the corpus is constructed at the sentence level, which limits the evaluation of document-level translation phenomena such as discourse coherence, context-dependent translation, and coreference resolution.

Our experiments focus on evaluating the effectiveness of COILD using representative multilingual neural machine translation models, namely IndicTrans2-Distilled and NLLB-200. While these models provide strong and reproducible baselines for multilingual MT, we do not evaluate recent large language models (LLMs) for multilingual MT. Finally, despite rigorous human translation and multi-stage quality assurance, subtle linguistic variations and regional preferences may still exist across some language pairs. We intend to continuously expand both the corpus and benchmark to improve language coverage, domain diversity, and linguistic representation.

\section*{Ethics Statement}

COILD was developed through an ethically responsible, human-centric data collection process. Source documents were obtained from publicly available or licensed Indian resources, with necessary permissions, and were screened to remove personally identifiable information, offensive content, and inappropriate material. Translations were produced manually by qualified native speakers who were recruited through a structured process and compensated for their work. All translations underwent multiple rounds of linguistic and domain-specific review to ensure accuracy, fluency, and consistency. Human evaluation was conducted independently by two native annotators for each translation direction. 

% Despite efforts to ensure regional diversity, some linguistic and regional variations may remain.

\section*{Acknowledgements}
The authors are grateful to the \textbf{Ministry of Electronics and Information Technology (MeitY), Government of India}, for supporting this work under \textbf{BHASHINI} initiative through the \textbf{COIL-D: Centre of Indian Language Data} (Project No. \textbf{SP/CSE/CIL/2024-25/1155}). We gratefully acknowledge all participating institutions and contributors whose collective efforts supported the development, translation, linguistic validation, technical implementation, and evaluation of the COIL-D corpus and benchmark. We sincerely acknowledge the contributions of the Principal Investigators, Co-Principal Investigators, project officers, research associates, language experts, translators, linguistic reviewers, domain experts, technical staff, administrative personnel, and other project contributors and freelancers.

\subsection*{Institute-wise Contributor}

\begin{itemize}

\item \textbf{IIT Patna:} Bengali, Santali, Maithili, and Odia. Contributors include Palash Gupta (Program Manager), Piyush Raj (Liaison Officer), Sauravi Haldar (JRA-Bengali), Ayodhya Murmu (JRA-Santali), Joysagar Murmu (JRA-Santali), Dr. Mukesh Kumar (JRA-Maithili) and Bigyan Ranjan Das (JRA-Odia).

\item \textbf{IIT Delhi:} Marathi, Gujarati, Kashmiri, and Telugu. Contributors include Sohini Mazumdar (Liaison Officer), Umesh Chandra Chaudhary (JRA-Hindi), Dr. Byrapuneni Bindu Madhavi (JRA-Telugu), Shailaja Gnanani (JRA-Telugu), Namburu Lakshmi Gayathri (JRA-Telugu), Vijay Wali (JRA-Kashmiri), Kishan A Pandya (JRA-Gujarati), Mahi Doshi (JRA-Gujarati), Premchand Babarao Nagdeote (JRA-Marathi) and Dr. Arfeen Zeeshan (SRA-Language).

\item \textbf{IIT Guwahati:} Assamese, Manipuri, Nepali, and Bodo. Contributors include Parismita Talukdar (JRA-Assamese), Pralay Kumar Boro (JRA-Bodo), Karishma Khakhlary (JRA-Bodo), Victoria Thangjam (JRA-Manipuri), Mayanglambam Yaiphabi Chanu (JRA-Manipuri), Dr. Shiben Sharma (JRA-Nepali), Tryfina Giri (JRA-Nepali) and Dr. Nongthombam Joychandra Singh (Project-Engineer).

\item \textbf{IIIT Delhi:} Sindhi, Urdu, and Konkani. Contributors include Rajnandini Pujari (Liaison Officer), Abdullah Mazhar (Technical JRA), Geeta Goklani (JRA-Sindhi), Izza Moin (JRA-Urdu), Nazia Akhtar (JRA-Urdu), Shamshul Arfin (JRA-Urdu), Vaishnavi Raikar (JRA-Konkani), and Siddhi Kerkar (JRA-Konkani).

\item \textbf{IGDTUW:} Dogri and Punjabi. Contributors include Ruchika Kaushik (Liaison Officer), Rahul Singh (JRA-Dogri), Kuldeep Kumar (JRA-Dogri), Manpreet Kaur (JRA-Punjabi), Tanya Jaiswal (JRA-Technical).

\item \textbf{MIT Manipal:} Tamil, Kannada, Telugu, and Malayalam. Contributors include Niveditha (Liaison Officer), Raksha Shetty (JRA-Kannada), Sona T (JRA-Malayalam), Shama Bhat (JRA-Kannada), Shruthi Naik (JRA-Kannada), Shrilatha (JRA-Kannada), and Harshavardhan Chowdary (JRA-Technical).

\end{itemize}

We sincerely appreciate the contributions of all individuals across these institutions whose expertise and support made the creation and validation of COILD possible.

\bibliography{custom}

\appendix

\section{Translation Guidelines}

To ensure the creation of high-quality parallel corpora, we developed standardised translation guidelines based on the COILD multilingual translation framework. The guidelines were designed under the respective language experts/faculties from central/state universities to produce translations that are linguistically accurate, culturally appropriate, and semantically faithful across all language pairs. The COILD has also organised the workshop/hands-on sessions with some language-specific colleges, including faculties and researchers, to obtain potential decisions on some clash points in translation rules and received fruitful decisions on that as well.

\subsection{Linguistic Analysis}

A contrastive linguistic analysis was conducted between the source and target Hindi-centric and Tamil-centric languages to identify structural differences in grammar, morphology, syntax, word order typology, agreement, honorific expressions, and lexical semantics aspects. The analysis helped identify language-specific challenges in translation and informed the development of the translation guidelines.

\subsection{Translation Rules and Language Conventions}

Based on linguistic analysis and the translation rules of the respective languages, the language-specific translation guidelines were prepared. These guidelines specify conventions for grammatical categories, named entities, technical terminology, foreign words, compound words, idiomatic expressions, abbreviations, numerals, punctuations and translation quality. Domain-specific recommendations were also provided for legal, healthcare, education, agriculture, governance, and science-related content.

\subsection{Translation and Post-editing}

% Native translators produce translations by following the predefined guidelines to preserve the semantic meaning while ensuring fluency and naturalness in the target language. Literal translations were avoided whenever they produced unnatural expressions, and culturally appropriate equivalents were preferred wherever possible. The core words of the respective languages from (Hindi-X and Tamil-Y and reverse) are strongly recommended to prefer while translation and post-editing with its original ethnolinguistic parameters. All the translations subsequently underwent review and post-editing by internal language-specific RAs to improve grammatical correctness, lexical consistency, and overall readability.

Native translators produced translations following predefined guidelines to preserve semantic meaning while ensuring fluency and naturalness in the target language. Literal translations were avoided when they resulted in unnatural expressions, and culturally appropriate equivalents were preferred wherever possible. Translators were encouraged to retain core lexical items and language-specific cultural elements, particularly for the Hindi-centric and Tamil-centric language pairs in both translation directions. All translations subsequently underwent review and post-editing by internal language-specific Research Associates to improve grammatical correctness, lexical consistency, and overall readability.

\subsection{Quality Assurance and Validation}

Each translated sentence was independently reviewed by native linguists and domain experts to ensure quality and naturalness in accordance with the guidelines. The review process focused on semantic accuracy, structural correctness, fluency, consistency of technical terminology, and cultural appropriateness. Any inconsistencies identified during the review were resolved through discussion before final acceptance. The final decisions, which were taken with the mutual consent under COILD have also been updated to the corpora creation guidelines over the period of time.

\subsection{Final Standardization}

The respective validated translations were standardised across all language pairs to ensure uniform formatting and consistent handling of punctuation, named entities, acronyms and abbreviations, numerals and technical terminology. We have also covered the updated replaced terms in the respective languages (vice-versa) which makes more or less significance presence in the translation world to map the updated footprint of translation parameters. The resulting multilingual corpus serves as a reliable resource for machine translation research and multilingual NLP applications.

% \section{Human Annotation Guidelines for Evaluation}
% Human evaluation compared the baseline and fine-tuned versions of IndicTrans2 and NLLB-200 using a five-point Likert scale across two quality dimensions: \textit{adequacy} and \textit{fluency}. Adequacy (1--5; 1=Poor, 5=Excellent) measures how faithfully the translation preserves the source meaning, while fluency evaluates grammatical correctness, readability, and naturalness.

% Our benchmark comprises 20 language pairs, corresponding to 40 translation directions. For each direction, 50 sentences were randomly sampled from each of eight evaluation domains, resulting in 400 evaluation sentences per direction. These sentences were translated using both baseline and fine-tuned versions of IndicTrans2 and NLLB-200.

\section{Human Annotation Guidelines for Evaluation}

Human evaluation compared the baseline and fine-tuned versions of IndicTrans2 and NLLB-200 using a five-point Likert scale across two quality dimensions: \textit{adequacy} and \textit{fluency}. Ratings ranged from 1 = Worst, 2 = Poor, 3 = Average, 4 = Good and 5 = Excellent. Adequacy measures how faithfully the translation preserves the meaning and intent of the source, including important information, terminology, and named entities. Fluency assesses grammatical correctness, readability, naturalness, and word choice in the target language. Annotators evaluated both criteria independently using the predefined guidelines.

The benchmark comprises 20 language pairs, corresponding to 40 translation directions. For each direction, 50 sentences were randomly sampled from each of eight evaluation domains, resulting in 400 evaluation sentences per direction. These sentences were translated using both baseline and fine-tuned versions of IndicTrans2 and NLLB-200. Each output was independently evaluated by two native-speaker annotators for adequacy and fluency. The final score for each criterion was calculated as the average of the ratings assigned by the two annotators.

\subsection{Annotation Procedure}

All translated outputs were independently received from the respective source evaluated by native linguists and respective language experts following a unified annotation protocol decided under COILD. Annotators assigned adequacy and fluency scores on a five-point Likert scale for every translation with the absolute understanding of translation correctness and respective language ethnological parameters. During annotation, the language related significance have been primarily taken care with the special attention, so that translation notion shouldn't shift to any other aspect in context to the meaning.

\subsection{Qualitative Remarks}

Along with the numerical scores, annotators recorded qualitative remarks for each evaluated translation. These remarks documented observed translation issues and highlighted improvements introduced by the fine-tuned models over the corresponding baseline systems, including better semantic preservation, improved fluency, more accurate terminology, enhanced grammatical correctness, and correction of translation errors. The same evaluation protocol was consistently applied across all 40 translation directions to ensure a fair and comprehensive comparison between the baseline and fine-tuned models.

\begin{table*}[t]
\centering
\scriptsize
\setlength{\tabcolsep}{3pt}
\renewcommand{\arraystretch}{1.12}

\begin{tabular}{lcccccc|cccccc}
\toprule

\rowcolor{headergray}

&
\multicolumn{6}{c|}{\textbf{IndicTrans2}}
&
\multicolumn{6}{c}{\textbf{NLLB}}
\\

\cmidrule(lr){2-7}
\cmidrule(lr){8-13}

\rowcolor{headergray}

\textbf{Direction}
&
\multicolumn{3}{c}{\textbf{Adequacy}}
&
\multicolumn{3}{c|}{\textbf{Fluency}}
&
\multicolumn{3}{c}{\textbf{Adequacy}}
&
\multicolumn{3}{c}{\textbf{Fluency}}
\\

\cmidrule(lr){2-4}
\cmidrule(lr){5-7}
\cmidrule(lr){8-10}
\cmidrule(lr){11-13}

\rowcolor{headergray}

&
\textbf{Baseline}
&
\textbf{Finetune}
&
\textbf{$\Delta$}
&
\textbf{Baseline}
&
\textbf{Finetune}
&
\textbf{$\Delta$}
&
\textbf{Baseline}
&
\textbf{Finetune}
&
\textbf{$\Delta$}
&
\textbf{Baseline}
&
\textbf{Finetune}
&
\textbf{$\Delta$}

\\
\midrule

Assamese $\rightarrow$ Hindi & 2.16 & 2.49 & \cellcolor{glight}0.33 & 3.01 & 3.35 & \cellcolor{glight}0.34 & 2.52 & 2.71 & \cellcolor{glight}0.19 & 3.42 & 3.54 & \cellcolor{glight}0.12 \\
Hindi $\rightarrow$ Assamese & 2.19 & 2.12 & \cellcolor{rlight}-0.07 & 3.03 & 2.82 & \cellcolor{rlight}-0.21 & 2.02 & 2.32 & \cellcolor{glight}0.30 & 2.66 & 2.80 & \cellcolor{glight}0.14 \\
Bengali $\rightarrow$ Hindi & 3.86 & 4.22 & \cellcolor{glight}0.36 & 4.15 & 4.44 & \cellcolor{glight}0.29 & 3.43 & 3.89 & \cellcolor{glight}0.47 & 3.52 & 4.02 & \cellcolor{glight}0.49 \\
Hindi $\rightarrow$ Bengali & 3.71 & 4.09 & \cellcolor{glight}0.38 & 3.99 & 4.27 & \cellcolor{glight}0.28 & 3.50 & 3.85 & \cellcolor{glight}0.35 & 3.56 & 4.02 & \cellcolor{glight}0.46 \\
Bodo $\rightarrow$ Hindi & 3.30 & 3.23 & \cellcolor{rlight}-0.06 & 3.35 & 3.34 & \cellcolor{rlight}-0.02 & -- & -- & -- & -- & -- & -- \\
Hindi $\rightarrow$ Bodo & 3.06 & 3.18 & \cellcolor{glight}0.13 & 3.06 & 3.18 & \cellcolor{glight}0.12 & -- & -- & -- & -- & -- & -- \\
Dogri $\rightarrow$ Hindi & 3.69 & 4.64 & \cellcolor{gmed}0.95 & 4.04 & 4.79 & \cellcolor{gmed}0.75 & -- & -- & -- & -- & -- & -- \\
Hindi $\rightarrow$ Dogri & 3.58 & 4.32 & \cellcolor{gmed}0.74 & 3.91 & 4.50 & \cellcolor{gmed}0.59 & -- & -- & -- & -- & -- & -- \\
Konkani $\rightarrow$ Hindi & 3.91 & 3.71 & \cellcolor{rlight}-0.20 & 3.92 & 3.58 & \cellcolor{rlight}-0.34 & 1.00 & 2.90 & \cellcolor{gdark}1.90 & 1.00 & 2.86 & \cellcolor{gdark}1.86 \\
Hindi $\rightarrow$ Konkani & 3.48 & 2.40 & \cellcolor{rdark}-1.08 & 3.14 & 2.44 & \cellcolor{rmed}-0.70 & 1.00 & 3.13 & \cellcolor{gdark}2.13 & 1.00 & 2.94 & \cellcolor{gdark}1.94 \\
Gujarati $\rightarrow$ Hindi & 3.99 & 4.36 & \cellcolor{glight}0.36 & 4.29 & 4.46 & \cellcolor{glight}0.17 & 3.90 & 4.38 & \cellcolor{glight}0.48 & 4.41 & 4.51 & \cellcolor{glight}0.09 \\
Hindi $\rightarrow$ Gujarati & 3.72 & 4.13 & \cellcolor{glight}0.40 & 4.07 & 4.26 & \cellcolor{glight}0.18 & 3.80 & 4.03 & \cellcolor{glight}0.23 & 4.17 & 4.37 & \cellcolor{glight}0.20 \\
Hindi $\rightarrow$ Kashmiri & 3.57 & 3.51 & \cellcolor{rlight}-0.06 & 3.59 & 3.56 & \cellcolor{rlight}-0.03 & 3.04 & 3.63 & \cellcolor{gmed}0.59 & 3.07 & 3.71 & \cellcolor{gmed}0.63 \\
Kashmiri $\rightarrow$ Hindi & 3.43 & 3.62 & \cellcolor{glight}0.20 & 3.76 & 3.89 & \cellcolor{glight}0.13 & 3.09 & 3.65 & \cellcolor{gmed}0.56 & 3.71 & 3.74 & \cellcolor{glight}0.03 \\
Hindi $\rightarrow$ Maithili & 2.34 & 3.80 & \cellcolor{gdark}1.46 & 2.25 & 3.74 & \cellcolor{gdark}1.49 & 2.79 & 3.91 & \cellcolor{gdark}1.12 & 2.39 & 3.74 & \cellcolor{gdark}1.35 \\
Maithili $\rightarrow$ Hindi & 2.59 & 3.86 & \cellcolor{gdark}1.27 & 2.52 & 3.82 & \cellcolor{gdark}1.30 & 3.07 & 3.83 & \cellcolor{gmed}0.77 & 2.68 & 3.67 & \cellcolor{gmed}0.99 \\
Hindi $\rightarrow$ Marathi & 3.74 & 4.09 & \cellcolor{glight}0.36 & 4.11 & 4.26 & \cellcolor{glight}0.15 & 3.17 & 3.79 & \cellcolor{gmed}0.62 & 3.67 & 4.06 & \cellcolor{glight}0.40 \\
Marathi $\rightarrow$ Hindi & 4.18 & 4.53 & \cellcolor{glight}0.35 & 4.48 & 4.70 & \cellcolor{glight}0.22 & 3.56 & 4.22 & \cellcolor{gmed}0.67 & 4.05 & 4.36 & \cellcolor{glight}0.30 \\
Hindi $\rightarrow$ Manipuri & 3.75 & 2.83 & \cellcolor{rmed}-0.92 & 3.80 & 3.01 & \cellcolor{rmed}-0.79 & -- & -- & -- & -- & -- & -- \\
Manipuri $\rightarrow$ Hindi & 3.92 & 4.29 & \cellcolor{glight}0.37 & 3.72 & 3.95 & \cellcolor{glight}0.23 & -- & -- & -- & -- & -- & -- \\
Hindi $\rightarrow$ Nepali & 4.01 & 4.27 & \cellcolor{glight}0.27 & 3.56 & 3.83 & \cellcolor{glight}0.27 & 4.01 & 4.34 & \cellcolor{glight}0.34 & 3.66 & 3.91 & \cellcolor{glight}0.25 \\
Nepali $\rightarrow$ Hindi & 4.46 & 4.71 & \cellcolor{glight}0.26 & 4.30 & 4.67 & \cellcolor{glight}0.38 & 4.13 & 4.62 & \cellcolor{glight}0.49 & 3.70 & 4.22 & \cellcolor{gmed}0.52 \\
Hindi $\rightarrow$ Odia & -- & -- & -- & -- & -- & -- & -- & -- & -- & -- & -- & -- \\
Odia $\rightarrow$ Hindi & -- & -- & -- & -- & -- & -- & -- & -- & -- & -- & -- & -- \\
Hindi $\rightarrow$ Punjabi & 3.53 & 3.96 & \cellcolor{glight}0.42 & 3.70 & 4.09 & \cellcolor{glight}0.39 & 3.49 & 4.16 & \cellcolor{gmed}0.68 & 3.53 & 4.08 & \cellcolor{gmed}0.55 \\
Punjabi $\rightarrow$ Hindi & 4.08 & 4.35 & \cellcolor{glight}0.27 & 4.32 & 4.48 & \cellcolor{glight}0.16 & 3.84 & 4.21 & \cellcolor{glight}0.36 & 4.17 & 4.30 & \cellcolor{glight}0.13 \\
Hindi $\rightarrow$ Santali & 2.22 & 3.18 & \cellcolor{gmed}0.96 & 2.69 & 3.49 & \cellcolor{gmed}0.80 & -- & -- & -- & -- & -- & -- \\
Santali $\rightarrow$ Hindi & 2.81 & 3.31 & \cellcolor{glight}0.49 & 3.21 & 3.57 & \cellcolor{glight}0.36 & -- & -- & -- & -- & -- & -- \\
Hindi $\rightarrow$ Sindhi & 3.18 & 3.26 & \cellcolor{glight}0.08 & 3.13 & 3.28 & \cellcolor{glight}0.15 & 1.00 & 3.41 & \cellcolor{gdark}2.41 & 1.00 & 3.25 & \cellcolor{gdark}2.25 \\
Sindhi $\rightarrow$ Hindi & 2.46 & 3.18 & \cellcolor{gmed}0.72 & 2.80 & 3.48 & \cellcolor{gmed}0.68 & 1.62 & 2.86 & \cellcolor{gdark}1.24 & 1.71 & 3.43 & \cellcolor{gdark}1.72 \\
Hindi $\rightarrow$ Telugu & 3.38 & 3.27 & \cellcolor{rlight}-0.12 & 3.68 & 3.53 & \cellcolor{rlight}-0.15 & 2.99 & 3.00 & \cellcolor{glight}0.01 & 3.60 & 3.47 & \cellcolor{rlight}-0.13 \\
Telugu $\rightarrow$ Hindi & 3.85 & 3.87 & \cellcolor{glight}0.03 & 3.98 & 3.93 & \cellcolor{rlight}-0.04 & 3.51 & 3.25 & \cellcolor{rlight}-0.26 & 3.76 & 3.55 & \cellcolor{rlight}-0.21 \\
Hindi $\rightarrow$ Urdu & 3.76 & 4.25 & \cellcolor{glight}0.49 & 4.10 & 4.24 & \cellcolor{glight}0.14 & 2.98 & 4.31 & \cellcolor{gdark}1.33 & 3.05 & 4.34 & \cellcolor{gdark}1.29 \\
Urdu $\rightarrow$ Hindi & 1.84 & 4.19 & \cellcolor{gdark}2.35 & 1.90 & 4.22 & \cellcolor{gdark}2.32 & 4.01 & 4.32 & \cellcolor{glight}0.31 & 4.12 & 4.30 & \cellcolor{glight}0.18 \\
Kannada $\rightarrow$ Tamil & 3.40 & 3.87 & \cellcolor{glight}0.47 & 3.40 & 3.82 & \cellcolor{glight}0.41 & 3.40 & 3.96 & \cellcolor{gmed}0.56 & 3.42 & 3.96 & \cellcolor{gmed}0.54 \\
Tamil $\rightarrow$ Kannada & 2.95 & 3.87 & \cellcolor{gmed}0.92 & 4.04 & 4.38 & \cellcolor{glight}0.34 & 2.81 & 4.06 & \cellcolor{gdark}1.26 & 3.99 & 4.56 & \cellcolor{gmed}0.57 \\
Malayalam $\rightarrow$ Tamil & 4.19 & 4.35 & \cellcolor{glight}0.16 & 4.45 & 4.56 & \cellcolor{glight}0.11 & 4.27 & 4.40 & \cellcolor{glight}0.13 & 4.48 & 4.57 & \cellcolor{glight}0.09 \\
Tamil $\rightarrow$ Malayalam & 3.15 & 3.68 & \cellcolor{gmed}0.52 & 3.93 & 4.67 & \cellcolor{gmed}0.74 & 3.09 & 3.87 & \cellcolor{gmed}0.78 & 3.86 & 4.18 & \cellcolor{glight}0.32 \\
Tamil $\rightarrow$ Telugu & 3.37 & 3.88 & \cellcolor{gmed}0.51 & 3.80 & 4.19 & \cellcolor{glight}0.39 & 2.91 & 4.00 & \cellcolor{gdark}1.09 & 3.21 & 4.06 & \cellcolor{gmed}0.85 \\
Telugu $\rightarrow$ Tamil & 3.97 & 4.29 & \cellcolor{glight}0.32 & 4.43 & 4.61 & \cellcolor{glight}0.18 & 3.61 & 4.17 & \cellcolor{gmed}0.56 & 4.24 & 4.51 & \cellcolor{glight}0.27 \\

\bottomrule
\end{tabular}

\vspace{1mm}
\caption{
Human evaluation results comparing the baseline and fine-tuned versions of IndicTrans2 and NLLB.
Adequacy and Fluency were rated by human annotators on a \textbf{5-point Likert scale} (1 = Worst, 2 = Poor, 3 = Average, 4 = Good, 5 = Excellent).
Positive $\Delta$ values indicate improvements after fine-tuning and are colour-coded to highlight the magnitude of performance changes after fine-tuning. Green shades indicate improvements, red shades indicate degradations. (Note: Odia Language human evaluation is underway.)
}
\label{tab:human_eval}
\end{table*}

\begin{figure*}[t]
\centering

\includegraphics[width=0.98\textwidth]{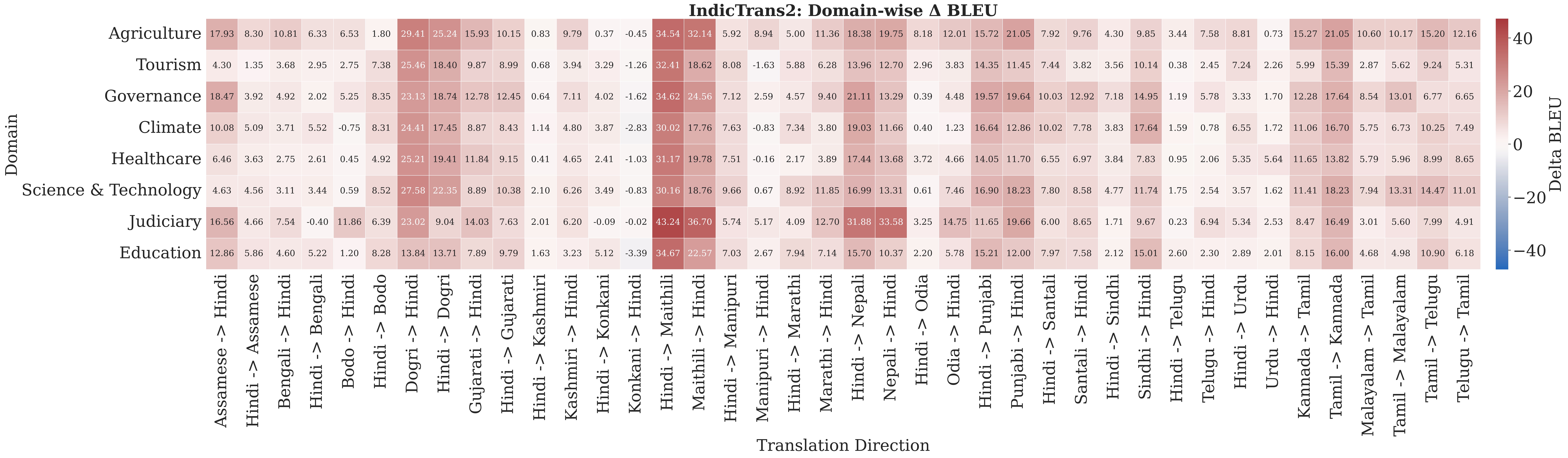}\\[2pt]
\includegraphics[width=0.98\textwidth]{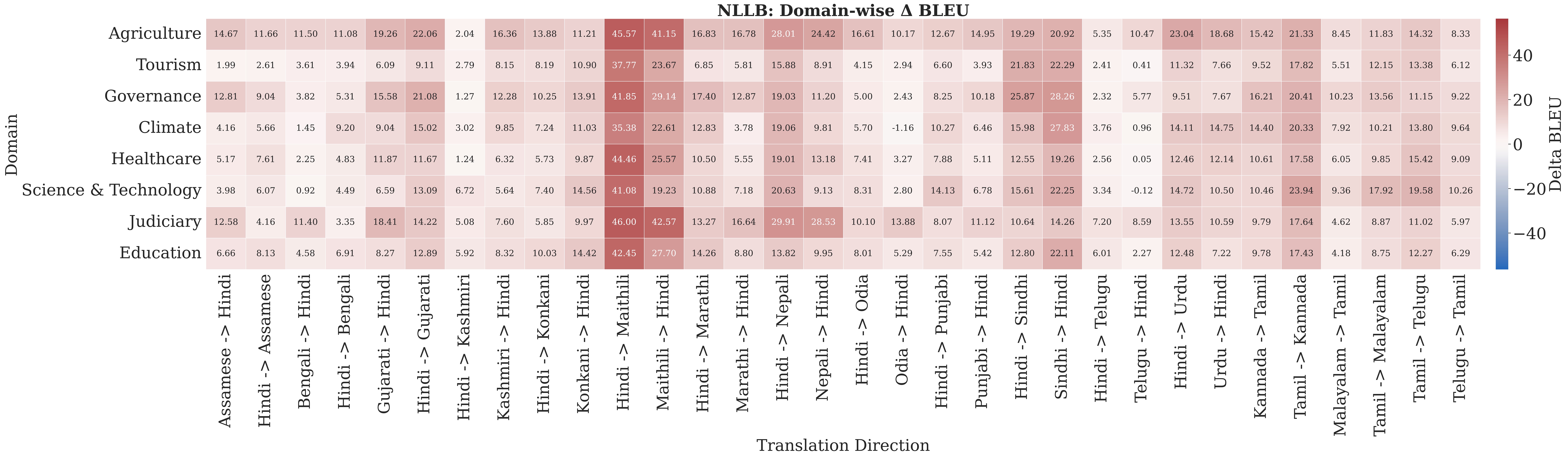}
\caption{Per-domain change in BLEU after finetuning on COILD, for IndicTrans2
(top) and NLLB-200 (bottom). Columns are translation directions, rows are
benchmark domains; cell values are $\Delta$BLEU (finetune $-$ baseline).}
\label{fig:domain-bleu}
\end{figure*}

\begin{figure*}[t]
\centering
\includegraphics[width=1\textwidth]{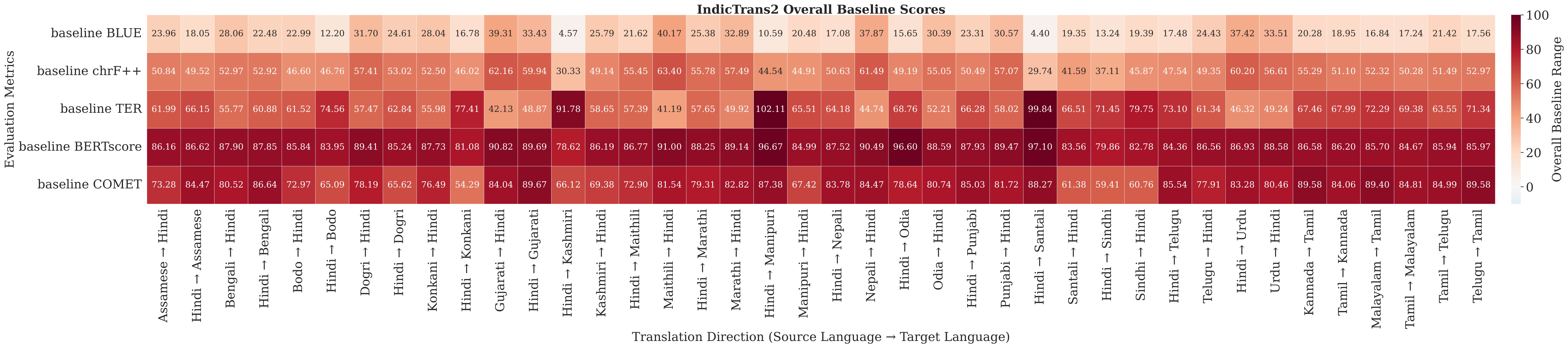}\\[2pt]
\includegraphics[width=1\textwidth]{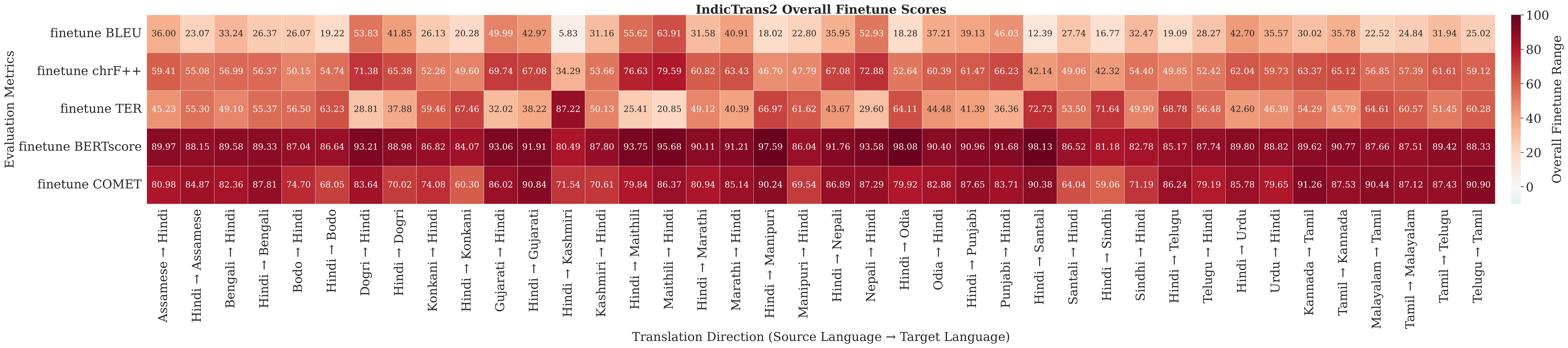}
\caption{Plot shows the overall baseline and finetune performance on the COILD Benchmark dataset for IndicTrans2.}
\label{fig:overall_baseline}
\end{figure*}

\begin{figure*}[t]
\centering
\includegraphics[width=1\textwidth]{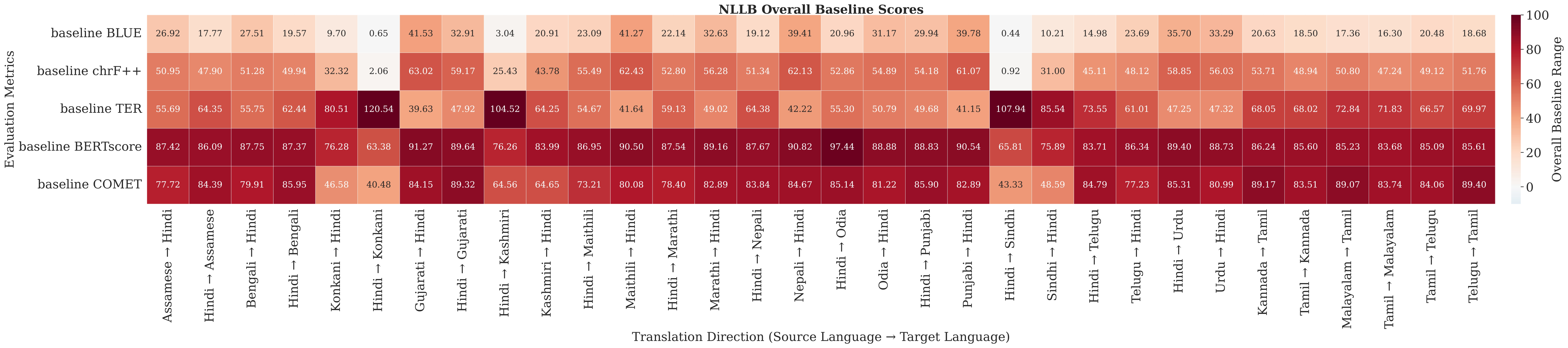}\\[2pt]
\includegraphics[width=1\textwidth]{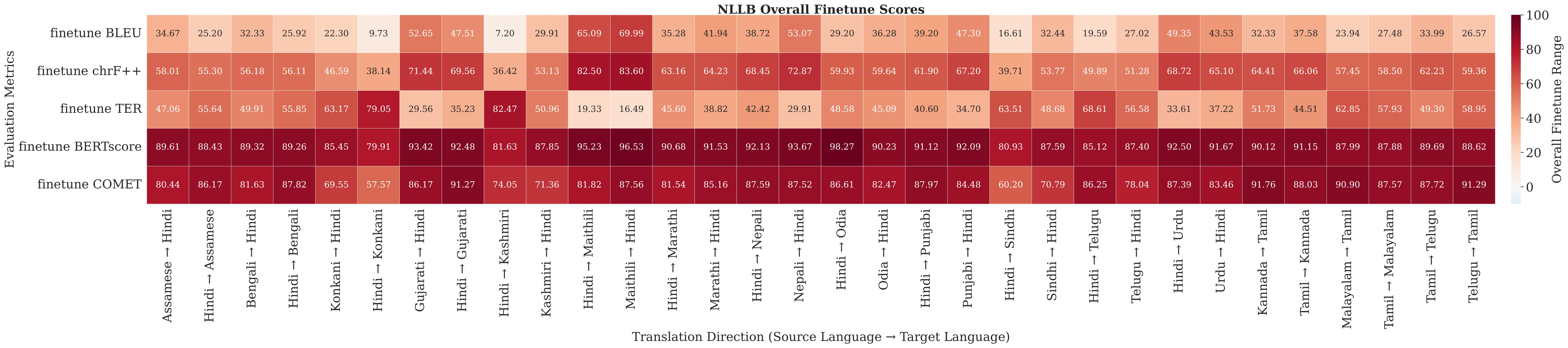}
\caption{Plot shows the overall baseline and finetune performance on the COILD Benchmark dataset for NLLB.}
\label{fig:overall_finetune}
\end{figure*}

\end{document}